\documentclass[final]{cvpr}

\usepackage{times}
\usepackage{epsfig}
\usepackage{graphicx}
\usepackage{amsmath}
\usepackage{amssymb}

\usepackage{xcolor, soul}
\sethlcolor{pink}

\usepackage{xparse, mathtools, array}
\DeclarePairedDelimiterX{\rvect}[1]{[}{]}{\,\makervect{#1}\,}
\ExplSyntaxOn
\NewDocumentCommand{\makervect}{m}
 {
  \seq_set_split:Nnn \l_tmpa_seq { , } { #1 }
  \begin{matrix}
  \seq_use:Nn \l_tmpa_seq { & }
  \end{matrix}
 }
\ExplSyntaxOff
\newcommand{\Transp}{\mathsf{T}}

\usepackage{amssymb}

\newcommand{\parsection}[1]{\vspace{2mm}\noindent\textbf{#1}~ }

\usepackage{booktabs}
\usepackage{multirow}

\usepackage{algorithm}
\usepackage[noend]{algpseudocode}
\makeatletter
\def\BState{\State\hskip-\ALG@thistlm}
\makeatother
\algnewcommand\Or{\textbf{or}}

\usepackage{amsmath,amsfonts,bm}

\def\eqref#1{equation~\ref{#1}}

\def\1{\bm{1}}

\DeclareMathAlphabet{\mathsfit}{\encodingdefault}{\sfdefault}{m}{sl}
\SetMathAlphabet{\mathsfit}{bold}{\encodingdefault}{\sfdefault}{bx}{n}

\DeclareMathOperator*{\argmax}{arg\,max}

\usepackage[caption=false]{subfig}

\makeatletter
\@namedef{ver@everyshi.sty}{}
\makeatother
\usepackage{tikz}

\usepackage{pgfplots}
    \usepgfplotslibrary{colorbrewer}
    \pgfplotsset{
        cycle list/Dark2,
        cycle multiindex* list={
            mark list*\nextlist
            Dark2\nextlist
        },
    }
\pgfplotsset{compat=1.14}

\usepackage[pagebackref=true,breaklinks=true,colorlinks,bookmarks=false]{hyperref}

\def\cvprPaperID{9} 
\def\confYear{CVPR 2021}
\begin{document}

\title{Accurate 3D Object Detection using Energy-Based Models}

\author{Fredrik K. Gustafsson$^{1}$
\and
Martin Danelljan$^{2}$
\and
Thomas B. Sch\"on$^{1}$ \vspace{1.0mm}
\and
$^{1}$Department of Information Technology, Uppsala University, Sweden
\and
$^{2}$Computer Vision Lab, ETH Z\"urich, Switzerland
}

\maketitle
\thispagestyle{empty}

\begin{abstract}
Accurate 3D object detection (3DOD) is crucial for safe navigation of complex environments by autonomous robots. Regressing accurate 3D bounding boxes in cluttered environments based on sparse LiDAR data is however a highly challenging problem. We address this task by exploring recent advances in conditional energy-based models (EBMs) for probabilistic regression. While methods employing EBMs for regression have demonstrated impressive performance on 2D object detection in images, these techniques are not directly applicable to 3D bounding boxes. In this work, we therefore design a differentiable pooling operator for 3D bounding boxes, serving as the core module of our EBM network. We further integrate this general approach into the state-of-the-art 3D object detector SA-SSD. On the KITTI dataset, our proposed approach consistently outperforms the SA-SSD baseline across all 3DOD metrics, demonstrating the potential of EBM-based regression for highly accurate 3DOD. Code is available at \url{https://github.com/fregu856/ebms_3dod}.
\end{abstract}

\section{Introduction}
\label{section:introduction}

\begin{figure}[t]
    \centering
    \includegraphics[width=1.0\linewidth]{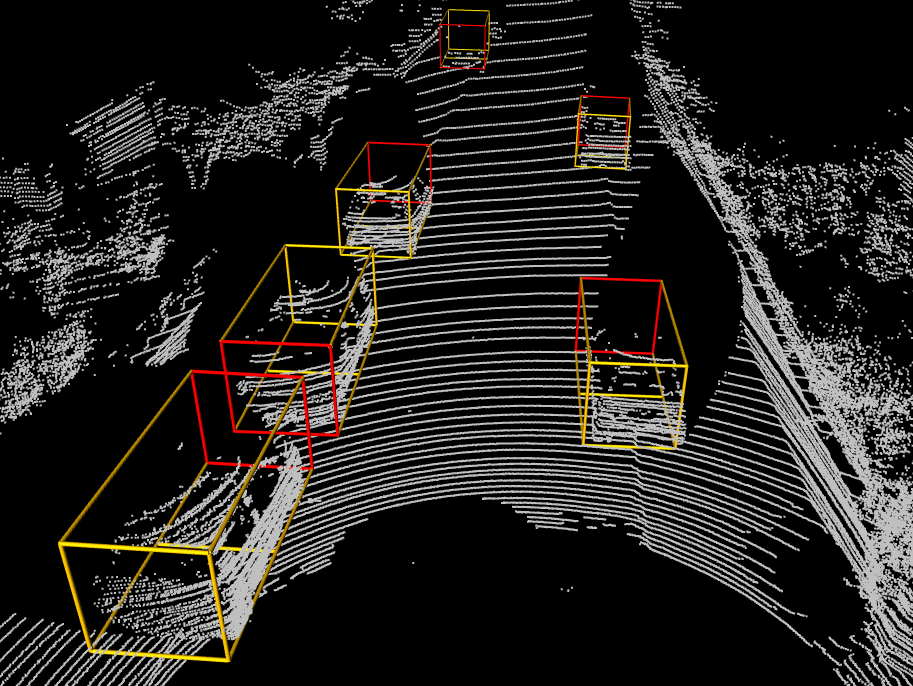}\vspace{0.0mm}
    \caption{We study how energy-based models (EBMs) can be applied to accurately regress 3D bounding boxes in 3DOD from LiDAR point clouds. Here, we visualize the output of our detector on a validation example from the KITTI~\cite{geiger2012we} dataset.}
    \label{fig:3dod}
\end{figure}

3D object detection (3DOD) is a key perception task for self-driving vehicles and other autonomous robots. 3DOD entails detecting various objects from sensor data, and estimating their size and position in the 3D world. Specifically, the goal of 3DOD is to place oriented 3D bounding boxes which tightly contain all surrounding objects of interest. See Figure~\ref{fig:3dod} for an example. These 3D bounding boxes then serve as input to important high-level tasks such as planning and collision avoidance. Accurate 3DOD is thus crucial for safe autonomous navigation of different complex environments.

In the automotive domain, 3DOD is usually performed from LiDAR point clouds \cite{yang2019std, shi2020points, shi2020pv}, images captured by vehicle-mounted cameras \cite{simonelli2019disentangling, chen2020monopair, shi2020distance}, or from a combination of both data modalities \cite{liang2019multi, liang2018deep, ku2018joint}. Radar sensors are sometimes also utilized \cite{meyer2019automotive, meyer2019deep, yang2020radarnet}. State-of-the-art 3D object detectors employ deep neural networks (DNNs) to learn powerful feature representations directly from this data \cite{shi2020pv, pang2020CLOCs, yoo20203d}. The 3DOD task is then commonly divided into two sub-tasks, in which anchor or proposal 3D bounding boxes are classified as either background or a specific class of object, and then regressed toward ground truth boxes \cite{zhou2018voxelnet, lang2019pointpillars, shi2019pointrcnn}.

\begin{figure*}[t]
    \centering
    \includegraphics[width=0.935\textwidth]{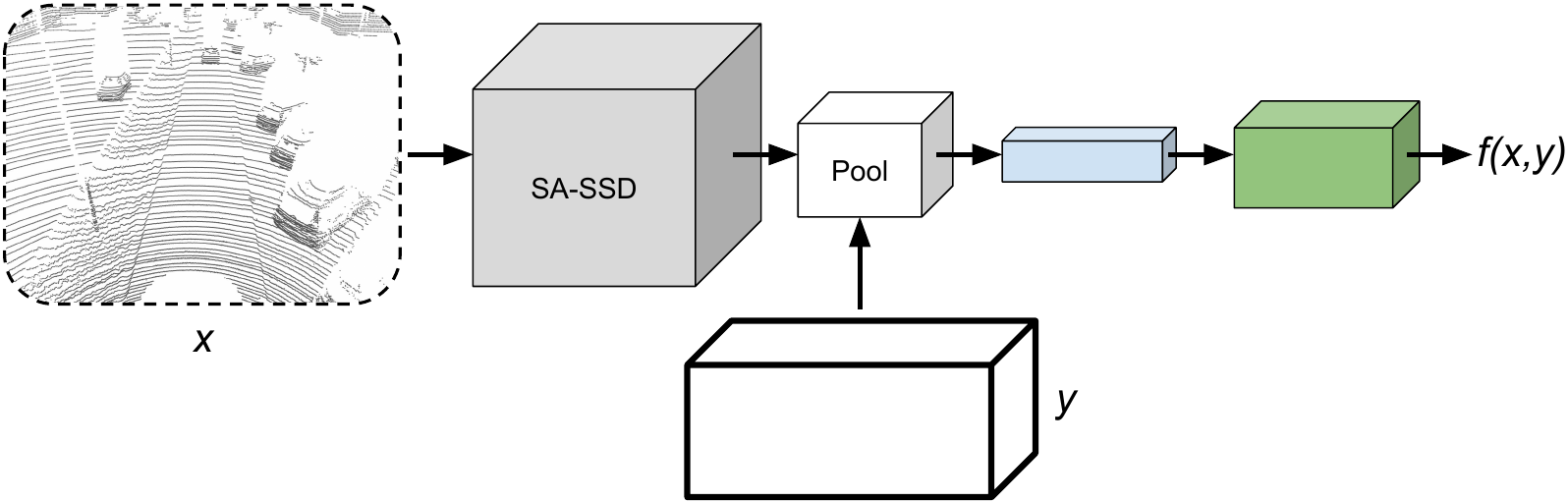}\vspace{0.5mm}
    \caption{An overview of our proposed approach, applying EBM-based regression to the task of 3D object detection. We integrate a conditional EBM $p(y | x; \theta) = e^{f_\theta(x, y)}/\int e^{f_\theta(x, \tilde{y})} d\tilde{y}$ into the state-of-the-art 3D object detector SA-SSD~\cite{he2020structure}. We achieve this by designing a differentiable pooling operator that, given a 3D bounding box $y$, extracts a feature vector from the SA-SSD output. This feature vector is then processed by three fully-connected layers, outputting the scalar energy $f_\theta(x, y) \in \mathbb{R}$.}\vspace{-1.5mm}
    \label{fig:overview}
\end{figure*}

In general, regression entails predicting a continuous target $y$ from an input $x$. This is a fundamental machine learning problem that can be addressed using a variety of different techniques \cite{lathuiliere2019comprehensive, gast2018lightweight, makansi2019overcoming, pan2018mean, Diaz_2019_CVPR}. Specifically in 3DOD, the 3D bounding box regression problem is usually addressed by letting a DNN directly predict a target bounding box $y$ for a given input $x$, and training the DNN by minimizing the $L^2$ or Huber loss~\cite{huber1964robust, zhou2018voxelnet, yang2019std, shi2020pv, he2020structure}. Alternatively, a probabilistic regression approach has also been employed. The conditional target density $p(y | x)$, i.e.\ the distribution for the target 3D bounding box $y$ given the input $x$, is then explicitly modelled using a DNN, which is trained by minimizing the associated negative log-likelihood. Previous work on 3DOD has mainly explored Gaussian models of $p(y | x)$~\cite{feng2018towards, feng2019leveraging, feng2019can, meyer2019lasernet}.

A Gaussian model is however fairly restrictive, limiting $p(y | x)$ to unimodal and symmetric distributions. Instead, recent work \cite{gustafsson2019learning, danelljan2020probabilistic, gustafsson2020train} has demonstrated that improved regression accuracy can be obtained on various tasks by employing energy-based models (EBMs)~\cite{lecun2006tutorial} to represent the conditional target density $p(y | x)$. Specifically, this approach entails modeling $p(y | x)$ with the conditional EBM $p(y | x; \theta) = e^{f_\theta(x, y)}/\int e^{f_\theta(x, \tilde{y})} d\tilde{y}$, and then using gradient ascent to maximize $p(y | x; \theta)$ w.r.t.\ $y$ at test-time. Since the EBM $p(y | x; \theta)$ is directly specified via the scalar function $f_\theta(x, y)$, which is defined using a DNN, it is a highly expressive model that puts minimal restricting assumptions on $p(y | x)$. Even potential multi-modality in the distribution $p(y | x)$ can therefore be learned directly from data. This EBM-based regression approach is thus an attractive alternative also for 3D bounding box regression, especially considering the impressive performance demonstrated on conventional 2D bounding box regression in images~\cite{gustafsson2019learning, danelljan2020probabilistic, gustafsson2020train}.

Extending the approach from 2D to 3D is however challenging. In particular, using gradient ascent to maximize the EBM $p(y | x; \theta)$ at test-time requires the scalar DNN output $f_\theta(x, y)$ to be differentiable w.r.t.\ the bounding box $y$. For 2D bounding boxes in images, this was achieved by applying a differentiable pooling operator \cite{jiang2018acquisition} on image features \cite{gustafsson2019learning, danelljan2020probabilistic, gustafsson2020train}, but this technique is not directly applicable to 3D bounding boxes. How EBM-based regression should be applied to 3DOD is thus currently an open question, which we set out to investigate in this work.

\parsection{Contributions}
We apply conditional EBMs $p(y | x; \theta)$ to the task of 3D bounding box regression, extending the recent EBM-based regression approach \cite{gustafsson2019learning, danelljan2020probabilistic, gustafsson2020train} from 2D to 3D object detection. This is achieved by adding an extra network branch to the state-of-the-art 3D object detector SA-SSD~\cite{he2020structure}, and designing a differentiable pooling operator for 3D bounding boxes $y$. We evaluate our proposed detector on the KITTI~\cite{geiger2012we} dataset and consistently outperform the SA-SSD baseline detector across all 3DOD metrics. Our work thus demonstrates the potential of EBM-based regression for highly accurate 3DOD.

\section{Energy-Based Models for Regression}
\label{section:ebms_regression}

EBMs were extensively studied by the machine learning community in the past \cite{lecun2006tutorial, teh2003energy, bengio2003neural, mnih2005learning, hinton2006unsupervised, osadchy2005synergistic}. In recent years they have also had a resurgence within the field of computer vision, frequently being employed for generative image modeling \cite{xie2016theory, gao2018learning, nijkamp2019learning, du2019implicit, Grathwohl2020Your, gao2020flow, pang2020learning, bao2020bi}. In comparison, the application of EBMs to regression problems has not been a particularly well-studied topic.  Very recent work~\cite{gustafsson2019learning, danelljan2020probabilistic, gustafsson2020train} has however demonstrated their efficacy on diverse computer vision regression tasks such as visual object tracking, head-pose estimation and age estimation.

In regression, the task is to learn to predict targets $y^\star \in \mathcal{Y}$ from inputs $x^\star \in \mathcal{X}$, given a training set $\mathcal{D}$ of i.i.d.\ input-target pairs, $\mathcal{D} = \{(x_i, y_i)\}_{i=1}^{N}$, $(x_i, y_i) \sim p(x, y)$. The input space $\mathcal{X}$ depends on the specific problem, but can e.g. correspond to the space of images or point clouds. The target space $\mathcal{Y}$ is continuous, $\mathcal{Y}=\mathbb{R}^K$ for some $K \geq 1$.

In EBM-based regression \cite{gustafsson2019learning, danelljan2020probabilistic, gustafsson2020train}, this task is addressed by modelling the distribution $p(y | x)$ of $y$ given $x$ with a conditional EBM $p(y | x; \theta)$, defined according to,
\begin{equation}
    p(y | x; \theta) = \frac{e^{f_{\theta}(x, y)}}{Z(x, \theta)}, \qquad Z(x, \theta) = \int e^{f_{\theta}(x, \tilde{y})} d\tilde{y}.
\label{eq:ebm_def}
\end{equation}
Here, $f_{\theta}: \mathcal{X} \times \mathcal{Y} \rightarrow \mathbb{R}$ is a DNN that maps any input-target pair $(x, y) \in \mathcal{X} \times \mathcal{Y}$ directly to a scalar $f_{\theta}(x, y) \in \mathbb{R}$, and $Z(x, \theta)$ is the input-dependent normalizing partition function. The DNN output $f_{\theta}(x, y)$ is interpreted as the (negative) energy of the distribution $p(y | x; \theta)$.

\subsection{Prediction}
\label{section:ebms_regression_pred}
At test-time, EBM-based regression entails predicting the most likely target under the model given an input $x^{\star}$, i.e. $y^\star = \argmax_y p(y | x^\star; \theta) = \argmax_y f_\theta(x^\star, y)$. In practice, $y^\star = \argmax_y f_\theta(x^\star, y)$ is approximated by refining an initial estimate $\hat{y}$ via $T$ steps of gradient ascent,
\begin{equation}
    y \gets y + \lambda \nabla_{y} f_{\theta}(x^\star, y),
\label{eq:gradient_asc}
\end{equation} 
thus finding a local maximum of $f_\theta(x^\star, y)$. Evaluation of the partition function $Z(x^\star, \theta)$ is therefore not required.

\subsection{Training}
\label{section:ebms_regression_train}
The DNN $f_\theta(x, y)$ that specifies the conditional EBM~(\ref{eq:ebm_def}) can be trained using various methods for fitting a density $p(y | x; \theta)$ to observed data $\{(x_i, y_i)\}_{i=1}^{N}$. Generally, the most straightforward such method is probably to minimize the negative log-likelihood $\mathcal{L}(\theta) = -\sum_{i=1}^{N} \log p(y_i | x_i; \theta)$, which for the EBM $p(y | x; \theta)$ is given by,  
\begin{equation}
    \mathcal{L}(\theta) = \sum_{i=1}^{N} \log \bigg( \int e^{f_{\theta}(x_i, y)} dy \bigg) - f_{\theta}(x_i, y_i).
\label{eq:ebm_nll}
\end{equation}
The integral in (\ref{eq:ebm_nll}) is however intractable, preventing exact evaluation of $\mathcal{L}(\theta)$. One possible solution to this problem is to approximate the intractable integral using importance sampling, as employed in \cite{gustafsson2019learning}. However, numerous alternative approaches also exist, including noise contrastive estimation (NCE) \cite{gutmann2010noise} and score matching \cite{hyvarinen2005estimation}. The problem of how EBMs should be trained specifically for regression was studied in detail in \cite{gustafsson2020train}, comparing six  methods on the task of 2D bounding box regression in images. From this comparison, \cite{gustafsson2020train} concluded that a simple extension of NCE should be considered the go-to training method.

NCE entails learning to discriminate between observed data examples and samples drawn from a noise distribution. NCE was adopted for EBM-based regression only recently in \cite{gustafsson2020train}, but has often been used to train EBMs for classification tasks in the past \cite{mnih2012fast, mikolov2013distributed, jozefowicz2016exploring, ma2018noise}. Recently, it has also become highly utilized within self-supervised representation learning \cite{hjelm2018learning, bachman2019learning, chen2020simple, han2020self}. Applying NCE to regression means training the DNN $f_\theta(x, y)$ by minimizing the loss,
\begin{equation}
\begin{gathered}
    J(\theta) = - \frac{1}{N} \sum_{i = 1}^{N} J_i(\theta),\\
    J_i(\theta)\!=\!\log\!\frac{\exp\!\big\{f_{\theta}(x_i, y_i^{(0)})\!-\!\log q(y_i^{(0)} | y_i)\!\big\}}{\sum\limits_{m=0}^{M} \exp\!\big\{f_{\theta}(x_i, y_i^{(m)})\!-\!\log q(y_i^{(m)} | y_i)\!\big\}},
\label{eq:nce_loss} 
\end{gathered}
\end{equation}
where $y_i^{(0)} \triangleq y_i$, and $\{y_i^{(m)}\}_{m=1}^{M}$ are $M$ samples drawn from a noise distribution $q(y|y_i)$ that depends on the true target~$y_i$. Effectively, $J(\theta)$ in (\ref{eq:nce_loss}) is the softmax cross-entropy loss for a classification problem with $M+1$ classes. A simple choice for $q(y|y_i)$ that was shown effective in \cite{gustafsson2020train} is setting $q$ to a mixture of $K$ Gaussians centered at $y_i$, 
\begin{equation}
    q(y|y_i) = \frac{1}{K} \sum_{k=1}^{K} \mathcal{N}(y; y_i, \sigma_{k}^{2}I),
\label{eq:nce_noise}
\end{equation}
where $K$ and the variances $\{\sigma^2_{k}\}_{k=1}^{K}$ are hyperparameters.

A simple extension to NCE, termed NCE+, was proposed and demonstrated to further improve the regression accuracy on certain tasks in \cite{gustafsson2020train}. The DNN $f_\theta$ is still trained by minimizing $J(\theta)$ in (\ref{eq:nce_loss}), but $y_i^{(0)}$ is now defined as $y_i^{(0)} \triangleq y_i + \nu_i$. The true target $y_i$ is thus perturbed with $\nu_i \sim q_{\beta}(y)$, where $q_{\beta}$ is a zero-centered and scaled version of $q(y|y_i)$ in (\ref{eq:nce_noise}), i.e. $q_{\beta}(y) = \frac{1}{K} \sum_{k=1}^{K} \mathcal{N}(y; 0, \beta \sigma_{k}^{2}I)$. NCE+ accounts for possible inaccuracies in the annotation process producing $y_i$, and can be understood as a direct generalization of NCE. In fact, NCE is recovered as a special case when $\beta \rightarrow 0$ in $q_{\beta}(y)$.

\section{Method}
\label{section:method}

We apply EBM-based regression to 3DOD by extending the state-of-the-art 3D object detector SA-SSD \cite{he2020structure} with a conditional EBM $p(y | x; \theta)$ (\ref{eq:ebm_def}). In Sec.~\ref{section:method_sassd}, we first provide necessary background on SA-SSD, including a description of its input and output data format. We then detail how the EBM $p(y | x; \theta)$ is defined, employing differentiable pooling of 3D bounding boxes $y$ and an added network branch, in Sec.~\ref{section:method_architecture}. Our approach for training $p(y | x; \theta)$ is based on NCE and further described in Sec.~\ref{section:method_training}. Lastly, our prediction strategy using gradient ascent is detailed in Sec.~\ref{section:method_prediction}.

\subsection{The SA-SSD 3D Object Detector}
\label{section:method_sassd}
SA-SSD \cite{he2020structure} takes a LiDAR point cloud of the scene as input $x$ and produces a set $\{d_i\}_{i=1}^D$ of $D$ detections. Each detection $d$ consists of a predicted 3D bounding box $y$, 
\begin{equation}
    y = \rvect{c_x, c_y, c_z, h, w, l, \phi}^{\Transp} \in \mathbb{R}^7,
\label{eq:y_def}
\end{equation}
and an associated classification confidence score $s \in (0, 1)$. In (\ref{eq:y_def}), $(c_x, c_y, c_z)$ is the 3D coordinate of the bounding box center, $(h, w, l)$ is the 3D bounding box size, and $\phi$ is the heading angle of the bounding box.  

The input LiDAR point cloud $x = \{(p_x^{(i)}, p_y^{(i)}, p_z^{(i)})\}_{i=1}^{n}$ of $n$ points is encoded into a sparse 3D tensor by means of voxelization. This tensor is then processed by a backbone network utilizing submanifold sparse 3D convolutional layers \cite{yan2018second, graham20183d}, producing a 3D feature tensor $h_1(x)$ of shape $W \times L \times H \times C$. A bird's eye view (BEV) feature representation of the scene is then created by flattening $h_1(x)$ into the 2D feature map $h_2(x)$ of shape $W \times L \times HC$. Then, $h_2(x)$ is further processed by six standard 2D convolutional layers, outputting the feature map $h_3(x)$ of shape $W \times L \times C'$. Finally, $h_3(x)$ is fed to a detection network, in which two $1 \times 1$ convolutions are applied. The first outputs classification confidence scores and the second outputs offsets for a $W \times L$ grid of anchor 3D bounding boxes.

The SA-SSD backbone and detection networks are trained by minimizing a weighted sum of multiple losses. The focal loss \cite{lin2017focal} is employed for the classification sub-task, and the Huber loss~\cite{huber1964robust} is used for the regression of anchor bounding box offsets. Additionally, SA-SSD employs two losses stemming from auxiliary tasks. By inverting the voxelization via interpolation, 3D feature tensors in the backbone network are represented as point-wise feature vectors. These are then utilized for point-wise foreground segmentation, i.e. predicting whether or not a point lies within any ground truth 3D bounding box, and point-wise center offset regression, i.e. predicting the offset from a foreground point to the center of its 3D bounding box.

\begin{figure}[t]
    \centering
    \includegraphics[width=0.425\linewidth]{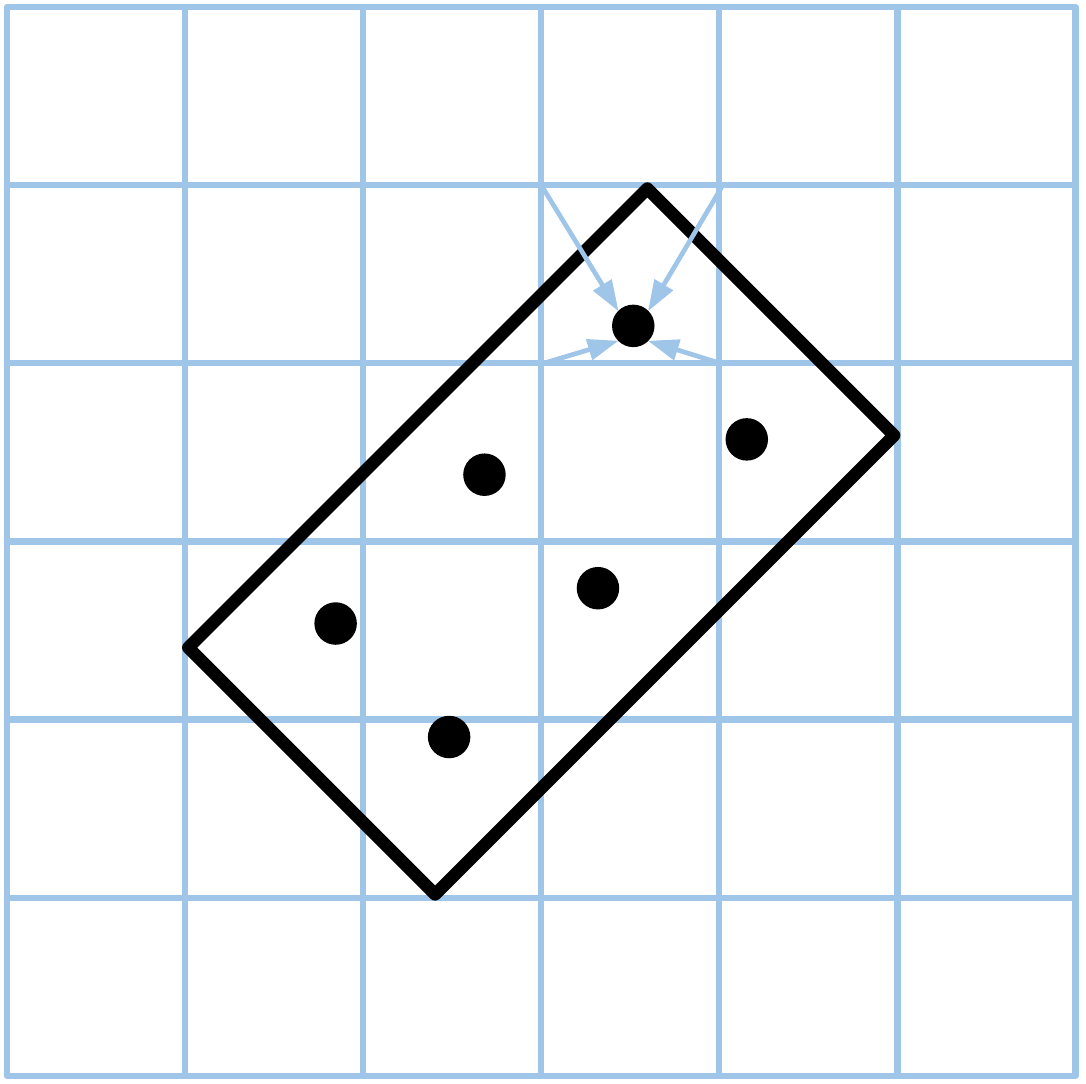}\vspace{0.5mm}
    \caption{Illustration of our modified variant of RoIAlign~\cite{He2017MaskR} for oriented 2D bounding boxes. In this example, the regular $W' \times L'$ grid is $2 \times 3$. Bilinear interpolation is used to extract a feature vector for each of the $W'L'$ grid points.}
    \label{fig:modified_roialign}
\end{figure}

\subsection{Conditional EBM Definition}
\label{section:method_architecture}
In this work, we extend the SA-SSD 3D object detector with a conditional EBM $p(y | x; \theta) = e^{f_\theta(x, y)}/\int e^{f_\theta(x, \tilde{y})} d\tilde{y}$, which is fully specified by the DNN $f_{\theta}$. To enable the use of gradient ascent at test time (Sec.~\ref{section:ebms_regression_pred}), the DNN must be designed such that its scalar output $f_{\theta}(x, y)$ is differentiable w.r.t. the 3D bounding box $y$ (\ref{eq:y_def}). To achieve this, we take inspiration from the recent work \cite{gustafsson2019learning, danelljan2020probabilistic, gustafsson2020train} applying EBM-based regression to 2D bounding box regression in images. Thus, we design a differentiable pooling operator that, for a given 3D bounding box $y$, extracts a feature vector from the SA-SSD backbone network output. This feature vector is then processed by an added network branch of fully-connected layers, outputting the energy value $f_{\theta}(x, y) \in \mathbb{R}$.

\parsection{Differentiable Pooling of 3D Bounding Boxes}
Various pooling operators for 3D bounding boxes $y$ (\ref{eq:y_def}) have been utilized for refining proposal bounding boxes in previous work \cite{shi2019pointrcnn, shi2020points, yang2019std, shi2020pv}, none of which are however differentiable w.r.t. the bounding box $y$. \cite{shi2019pointrcnn} extracts all points in the point cloud $x = \{(p_x^{(i)}, p_y^{(i)}, p_z^{(i)})\}_{i=1}^{n}$ which lie within a given box $y$, and then processes the associated point-wise features to extract a feature vector for $y$. This operator is however not differentiable w.r.t.\ $y$, due to the required discrete assessment of whether a point $(p_x^{(i)}, p_y^{(i)}, p_z^{(i)})$ lies within the 3D bounding box $y$ or not. \cite{shi2020points} instead divides the box $y$ into a 3D grid and extracts all points which lie within each grid cell. By also encoding which grid cells are empty, this pooling operator better captures geometric information. Because of the discrete extraction of points for each grid cell, it is however still not differentiable w.r.t.\ the 3D bounding box $y$. For similar reasons, the pooling operators utilized in \cite{yang2019std, shi2020pv}, which capture even richer contextual information, are not differentiable w.r.t.\ $y$ either.

Instead, we utilize the 2D feature map $h_3(x)$ of shape $W \times L \times C'$ that is produced by the SA-SSD backbone network. This is a compact yet powerful BEV feature representation of the scene. Specifically, we extract a feature vector $h_4(x, y^{\mathrm{BEV}})$ by pooling $h_3(x)$ with $y^{\mathrm{BEV}}$,
\begin{equation}
    y^{\mathrm{BEV}} = \rvect{c_x, c_y, w, l, \phi}^{\Transp} \in \mathbb{R}^5,
\label{eq:ybev_def}
\end{equation}
which is the BEV version of the 3D bounding box $y$~(\ref{eq:y_def}). Since $y^{\mathrm{BEV}}$ is an oriented 2D bounding box and not necessarily axis-aligned, we can not directly apply standard 2D bounding box pooling operators \cite{Girshick2015FastR, He2017MaskR, jiang2018acquisition}. Instead we employ a modified variant of RoIAlign~\cite{He2017MaskR}, which entails dividing $y^{\mathrm{BEV}}$ into a regular $W' \times L'$ grid, and extracting a feature vector $g \in \mathbb{R^{C'}}$ in each grid point via bilinear interpolation of $h_3(x)$. See Figure~\ref{fig:modified_roialign} for an illustration. This operation results in a 2D feature map of shape $W' \times L' \times C'$, which we then flatten to obtain the feature vector $h_4(x, y^{\mathrm{BEV}}) \in \mathbb{R}^{W'L'C'}$. By flattening the feature map instead of e.g. averaging over it, more information is preserved in $h_4(x, y^{\mathrm{BEV}})$. It can thus be used to discriminate between a given box and the same box rotated $\pi$ rad.

\begin{figure}[t]
    \centering
    \includegraphics[width=1.0\linewidth]{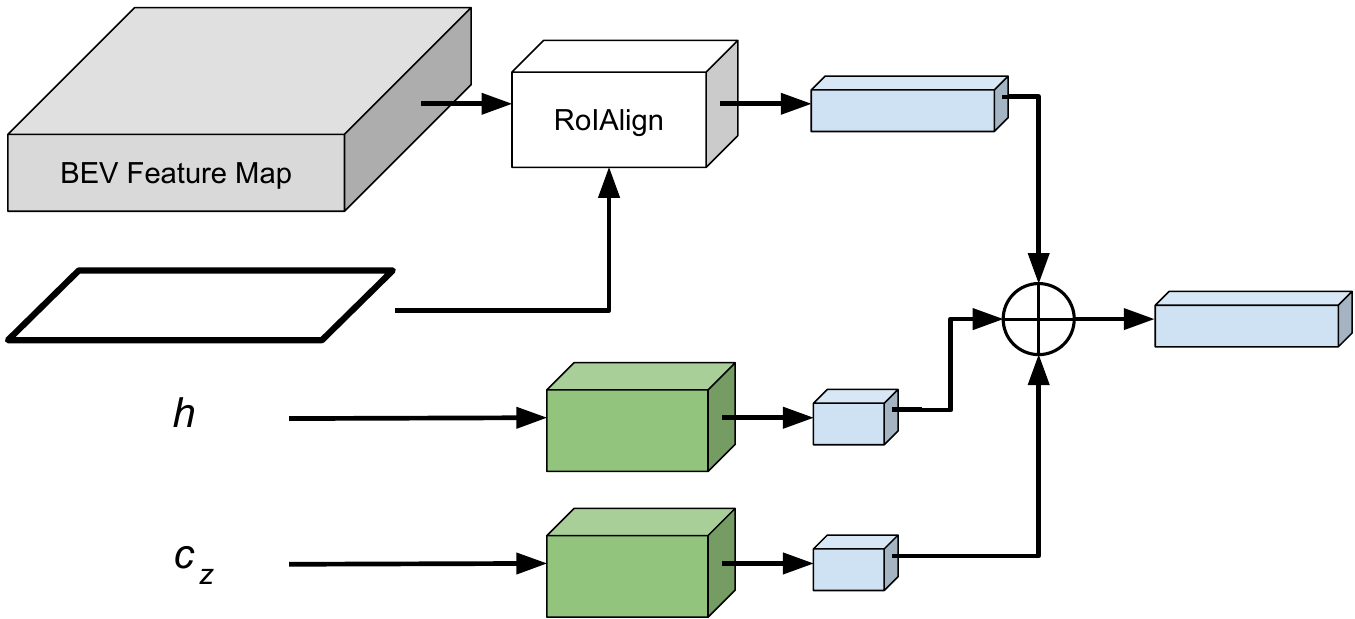}\vspace{0.5mm}
    \caption{Detailed illustration of the proposed differentiable pooling operation from 3D bounding box $y$ (\ref{eq:y_def}) to feature vector $h_5(x, y)$ (\ref{eq:featurevec_def}). The BEV version of $y$ is pooled with the BEV feature map produced by SA-SSD. The $z$ coordinate $c_z$ and height $h$ of the box $y$ are processed by two fully-connected layers.}
    \label{fig:3dbbox_pooling}
\end{figure}

This pooling operation is differentiable w.r.t.\ $y^{\mathrm{BEV}}$, but the extracted feature vector $h_4(x, y^{\mathrm{BEV}}) \in \mathbb{R}^{W'L'C'}$ is of course only a function of $y^{\mathrm{BEV}}$~(\ref{eq:ybev_def}), not of the full 3D bounding box $y$~(\ref{eq:y_def}). Using gradient ascent at test-time would thus not update the $z$ coordinate $c_z$ or height $h$ of the bounding box $y$. To resolve this, we take inspiration from the architecture used for EBM-based age estimation \cite{gustafsson2019learning}. We thus process $c_z \in \mathbb{R}$ and $h \in \mathbb{R}$ by two small fully-connected layers, generating feature vectors $g_{c_z} \in \mathbb{R}^{C''}$ and $g_h \in \mathbb{R}^{C''}$. Finally, we concatenate the three vectors to obtain $h_5(x, y)$, 
\begin{equation}
    h_5(x, y) = h_4(x, y^{\mathrm{BEV}}) \oplus g_{c_z} \oplus g_h \in \mathbb{R}^{W'L'C' + 2C''},
\label{eq:featurevec_def}
\end{equation}
where $\oplus$ indicates vector concatenation. The complete pooling operation from 3D bounding box $y$ to feature vector $h_5(x, y)$ is illustrated in Figure~\ref{fig:3dbbox_pooling}.

\parsection{Energy Prediction Branch}
Following \cite{gustafsson2019learning, danelljan2020probabilistic, gustafsson2020train}, we add an extra network branch onto SA-SSD for processing the extracted feature vector. The network branch consists of three fully-connected layers. It takes the feature vector $h_5(x, y) \in \mathbb{R}^{W'L'C' + 2C''}$ as input and outputs the scalar energy $f_{\theta}(x, y) \in \mathbb{R}$, thus fully specifying the conditional EBM $p(y | x; \theta)$ (\ref{eq:ebm_def}). The complete architecture of $p(y | x; \theta)$ is illustrated in Figure~\ref{fig:overview}.

\subsection{Detector Training}
\label{section:method_training}
Following the work on EBM-based 2D object detection \cite{gustafsson2019learning, gustafsson2020train}, the extra fully-connected layers described in Sec~\ref{section:method_architecture} are added onto a pre-trained and fixed SA-SSD detector. The parameters $\theta$ in $f_\theta(x, y)$ thus only stem from these added fully-connected layers, and the SA-SSD backbone and detection networks are kept fixed during training of the DNN~$f_\theta$. To train $f_\theta$, we use NCE as described in Sec~\ref{section:ebms_regression_train}. We employ the same training parameters (batch size, data augmentation etc.) as for SA-SSD \cite{he2020structure}, only replacing the original detector loss with the NCE loss (\ref{eq:nce_loss}).

\begin{algorithm}
\caption{Gradient-based refinement.}
\label{algo:prediction}
\textbf{Input:} $x^\star$, $\{\hat{y}_i\}_{i=1}^D$, $T$, $\lambda$, $\eta$.
\begin{algorithmic}[1]
    \For{\texttt{$i = 1, \dots, D$}}
        \State $y \gets \hat{y}_i$.
        \For{\texttt{$t = 1, \dots, T$}}
            \State \texttt{PrevValue} $\gets$ $f_{\theta}(x^\star, y)$.
            \State $\Tilde{y} \gets y + \lambda \nabla_{y} f_{\theta}(x^\star, y)$.
            \State \texttt{NewValue} $\gets$ $f_{\theta}(x^\star, \Tilde{y})$.
            \If { $\texttt{NewValue} > \texttt{PrevValue}$}
                \State $y \gets \Tilde{y}$.
            \Else
                \State $\lambda \gets \eta \lambda$.
            \EndIf
        \EndFor
        \State $y_i \gets y$.
    \EndFor
    \State \textbf{Return} $\{y_i\}_{i=1}^D$.
\end{algorithmic}
\end{algorithm}

\subsection{Detector Inference}
\label{section:method_prediction}
At test-time, the input LiDAR point cloud $x^\star$ is first processed by the SA-SSD detector. SA-SSD outputs the 2D feature map $h_3(x^\star)$ and a set $\{(\hat{y}_i, s_i)\}_{i=1}^D$ of $D$ detections, where $\hat{y}_i$ is a 3D bounding box (\ref{eq:y_def}) and $s_i$ is the associated classification confidence score. We then take all bounding boxes $\{\hat{y}_i\}_{i=1}^D$ as initial estimates and refine these via $T$ steps of gradient ascent (Sec~\ref{section:ebms_regression_pred}), producing $\{y_i\}_{i=1}^D$. The initial 3D bounding boxes $\{\hat{y}_i\}_{i=1}^D$ are thus refined by being moved toward different local maxima of $f_\theta(x^\star, y)$. The refined boxes $\{y_i\}_{i=1}^D$ are finally combined with the original confidence scores, returning the detections $\{(y_i, s_i)\}_{i=1}^D$.

This gradient-based refinement of the detections produced by SA-SSD of course lowers the detector inference speed somewhat. The point cloud $x^\star$ is however still processed by SA-SSD only once, and the scalar $f_\theta(x^\star, y)$ is extracted from $h_3(x^\star)$ using an efficient pooling operator and just a few fully-connected layers. Moreover, the gradient $\nabla_{y} f_{\theta}(x^\star, y)$ can be efficiently evaluated using automatic differentiation. The complete refinement procedure is detailed in Algorithm~\ref{algo:prediction}, where $\lambda$ denotes the gradient ascent step-length, $\eta$ is a decay of the step-length, and the $\texttt{NewValue} > \texttt{PrevValue}$ check ensures that $f_\theta(x^\star, y)$ is never decreased.

\section{Experiments}
\label{section:experiments}

We evaluate our EBM-based 3DOD approach on the KITTI 3DOD dataset \cite{geiger2012we} and compare it with the SA-SSD~\cite{he2020structure} baseline and other state-of-the-art methods. Our detector is implemented in PyTorch~\cite{paszke2019pytorch}. Training and inference code is publicly available.

\begin{table}[t]
\centering
\resizebox{1.0\linewidth}{!}{%
	\begin{tabular}{@{~}l@{~}|@{~}c@{~}c@{~}c@{~}|@{~}c@{~}c@{~}c@{~}}
\toprule
 & &3D @ 0.7 & & &BEV @ 0.7 &\\
 &Easy &Moderate &Hard &Easy &Moderate &Hard\\
\midrule
Part-A$^2$~\cite{shi2020points} &87.81 &78.49 &73.51    &91.70 &87.79 &84.61\\
SERCNN~\cite{zhou2020joint} &87.74 &78.96 &74.30    &94.11 &88.10 &83.43\\
EPNet~\cite{huang2020epnet} &89.81 &79.28 &74.59    &94.22 &88.47 &83.69\\
Point-GNN~\cite{shi2020point} &88.33 &79.47 &72.29    &93.11 &89.17 &83.90\\
3DSSD~\cite{yang20203dssd} &88.36 &79.57 &74.55    &92.66 &89.02 &85.86\\
STD~\cite{yang2019std} &87.95 &79.71 &75.09    &94.74 &89.19 &86.42\\
SA-SSD~\cite{he2020structure} &88.75 &79.79 &74.16 &95.03 &\textbf{91.03} &85.96\\
3D-CVF~\cite{yoo20203d} &89.20 &80.05 &73.11  &93.52 &89.56 &82.45\\
CLOCs-PVCas~\cite{pang2020CLOCs} &88.94 &80.67 &\textbf{77.15} &93.05 &89.80 &\textbf{86.57}\\
PV-RCNN~\cite{shi2020pv}     &90.25 &\textbf{81.43} &76.82 &94.98 &90.65 &86.14\\
\midrule
\midrule
SA-SSD            &88.80 &79.52 &72.30 &95.44 &89.63 &84.34\\
\textbf{SA-SSD+EBM} &\textbf{91.05} &80.12 &72.78 &\textbf{95.64} &89.86 &84.56\\
\textit{Rel. Improvement}            &+2.53\% &+0.75\% &+0.66\% &+0.21\% &+0.26\% &+0.26\%\\
\bottomrule
\end{tabular}
}
\vspace{0.0mm}
\caption{Results on KITTI test in terms of 3D and BEV AP. Our SA-SSD+EBM detector consistently outperforms the SA-SSD baseline, and achieves highly competitive performance also compared to other state-of-the-art methods.}
\label{tab:kitti_test}
\end{table}


\subsection{Dataset} 
KITTI~\cite{geiger2012we} is the most commonly used dataset for automotive 3DOD. It contains 7\thinspace481 examples for training, and 7\thinspace518 \textit{test} examples without publicly available ground truth annotations. Following common practice \cite{he2020structure, shi2020pv}, the training examples are further divided into \textit{train} (3\thinspace712 examples) and \textit{val} (3\thinspace769 examples) splits. We train models exclusively on the \textit{train} split and set hyperparameters using the \textit{val} split. We report results both on \textit{val}, and on the \textit{test} split by submitting detections to the KITTI benchmark server. Following SA-SSD, we conduct experiments only on the car object class.

\parsection{Evaluation Metrics}
On the KITTI benchmark server, models are evaluated in terms of average precision (AP) in both 3D and BEV. It considers three different difficulty levels (easy, moderate and hard), based on how far away and occluded objects are. AP is the area under the precision-recall curve, where a predicted bounding box is considered a true positive if its 3D/BEV IoU with a ground truth box exceeds a certain threshold. For cars, the threshold is set to $0.7$ on the KITTI benchmark. Two predicted boxes with IoU of, e.g., $0.71$ and $0.99$ thus have identical effect on this metric. Since our main goal is to improve the accuracy of all predicted bounding boxes, we also report the AP for higher thresholds $\{0.75, 0.8, 0.85, 0.9\}$ on the \textit{val} split. All reported AP values are computed using 40 recall positions.

\begin{table}[t]
\centering
\resizebox{1.0\linewidth}{!}{%
	\begin{tabular}{@{~}l@{~}|@{~}c@{~}c@{~}c@{~}|@{~}c@{~}c@{~}c@{~}}
\toprule
 & &3D @ 0.7 & & &BEV @ 0.7 &\\
 &Easy &Moderate &Hard &Easy &Moderate &Hard\\
\midrule
SA-SSD~\cite{he2020structure} &93.23 &84.30 &81.36 &- &- &-\\
CLOCs-PVCas~\cite{pang2020CLOCs} &92.78 &85.94 &\textbf{83.25} &93.48 &91.98 &89.48\\
PV-RCNN~\cite{shi2020pv}     &92.57 &84.83 &82.69 &95.76 &91.11 &88.93\\
\midrule
\midrule
SA-SSD            &93.14 &84.65 &81.86 &96.56 &92.84 &90.36\\
\textbf{SA-SSD+EBM} &\textbf{95.45} &\textbf{86.83} &82.23 &\textbf{96.60} &\textbf{92.92} &\textbf{90.43}\\
\textit{Rel. Improvement}            &+2.48\% &+2.58\% &+0.45\% &+0.04\% &+0.09\% &+0.08\%\\
\bottomrule
\end{tabular}
}
\vspace{0.0mm}
\caption{Results on KITTI val in terms of 3D and BEV AP. Our proposed detector consistently outperforms the SA-SSD baseline, and sets a new state-of-the-art for all but one of the metrics.}
\label{tab:kitti_val}
\end{table}

\begin{table*}[t]
\centering
\resizebox{1.0\linewidth}{!}{%
	\begin{tabular}{@{~}l@{~}|@{~}c@{~}c@{~}c@{~}|@{~}c@{~}c@{~}c@{~}|@{~}c@{~}c@{~}c@{~}|@{~}c@{~}c@{~}c@{~}}
\toprule
 & &3D @ 0.75 & & &3D @ 0.8 & & &3D @ 0.85 & & &3D @ 0.9 &\\
 &Easy &Moderate &Hard &Easy &Moderate &Hard &Easy &Moderate &Hard &Easy &Moderate &Hard\\
\midrule
SA-SSD            &84.48 &73.91 &70.99 &60.89 &50.08 &47.37 &24.29 &19.58 &18.05 &2.06 &1.58 &1.33\\
\textbf{SA-SSD+EBM} &87.85 &74.96 &71.95 &66.70 &54.32 &51.36 &31.02 &23.91 &21.95 &3.45 &2.74 &2.26\\
\textit{Rel. Improvement}            &+3.99\% &+1.42\% &+1.35\% &+9.54\% &+8.47\% &+8.42\% &+27.7\% &+22.1\% &+21.6\% &+67.5\% &+73.4\% &+69.9\%\\

\midrule
\midrule

 & &BEV @ 0.75 & & &BEV @ 0.8 & & &BEV @ 0.85 & & &BEV @ 0.9 &\\
 &Easy &Moderate &Hard &Easy &Moderate &Hard &Easy &Moderate &Hard &Easy &Moderate &Hard\\
\midrule
SA-SSD            &95.41 &87.47 &84.79  &87.12 &79.07 &74.65  &61.53 &54.15 &50.39  &17.48 &15.71 &14.58\\
\textbf{SA-SSD+EBM} &95.47 &87.54 &84.88  &88.31 &80.06 &77.25  &68.40 &58.62 &54.48  &26.60 &22.03 &19.48\\
\textit{Rel. Improvement}            &+0.06\% &+0.08\% &+0.11\%  &+1.37\% &+1.25\% &+3.48\%  &+11.2\% &+8.25\% &+8.12\%  &+52.2\% &+40.2\% &+33.6\%\\
\bottomrule
\end{tabular}
}
\vspace{0.0mm}
\caption{Results on KITTI val in terms of 3D and BEV AP for higher thresholds $\{0.75, 0.8, 0.85, 0.9\}$. Our SA-SSD+EBM detector consistently outperforms the SA-SSD baseline across all metrics, and the relative improvement increases with the AP threshold.}
\label{tab:kitti_val_3d}
\end{table*}

\subsection{Implementation Details}
We utilize the open-source implementation and pre-trained model provided\footnote{\url{https://github.com/skyhehe123/SA-SSD}} by the SA-SSD authors. The feature map $h_3(x)$ that is produced by the backbone network is of shape $200 \times 176 \times 256$. We divide each $y^{\mathrm{BEV}}$ (\ref{eq:ybev_def}) into a regular $4 \times 7$ grid, meaning that the feature vector $h_4(x, y^{\mathrm{BEV}}) \in \mathbb{R}^{7168}$. We process $c_z \in \mathbb{R}$ and $h \in \mathbb{R}$ with separate fully-connected layers (dimensions: $1 \rightarrow 16$, $16 \rightarrow 16$), generating $g_{c_z} \in \mathbb{R}^{16}$ and $g_h \in \mathbb{R}^{16}$. After concatenation, we thus obtain $h_5(x, y) \in \mathbb{R}^{7200}$. Finally, $h_5(x, y)$ is processed by three fully-connected layers of dimensions $7200 \rightarrow 1024$, $1024 \rightarrow 1024$, $1024 \rightarrow 1$. To train the DNN $f_\theta(x, y)$, i.e. the added fully-connected layers, we just replace the original detector loss with the NCE loss (Sec.~\ref{section:method_training}). We also considered NCE+ with $\beta > 0$, but saw no clear improvements over NCE. We hypothesize this is because there is less inherent ambiguity in the annotation process of 3D bounding boxes than of 2D bounding boxes in images. As in \cite{gustafsson2020train}, we set $K=3$ with $\sigma_1 = \sigma_3/4$, $\sigma_2 = \sigma_3/2$ for the noise distribution $q(y|y_i)$ (\ref{eq:nce_noise}). After ablation, optimizing 3D AP (moderate difficulty) on the \textit{val} split, we set $\sigma_3$ differently for different components of the 3D box $y$ (\ref{eq:y_def}). Specifically, $\sigma_3 = 0.25$ for $(c_x, c_y)$, $\sigma_3 = 0.125$ for $(c_z, h, w, l)$ and $\sigma_3 = 0.0625$ for $\phi$. Following \cite{gustafsson2019learning, gustafsson2020train}, we also set $T = 10$ and $\eta = 0.5$ for gradient-based refinement (Algorithm~\ref{algo:prediction}). The step-length $\lambda = 0.0002$ was selected based on ablation.

\subsection{3DOD Results on KITTI}
Results on KITTI \textit{test} in terms of 3D and BEV AP ($0.7$ threshold) are found in Table~\ref{tab:kitti_test}. We mainly compare the performance of our EBM-based 3D object detector (SA-SSD+EBM) to the pre-trained SA-SSD it extends, and include other state-of-the-art detectors for reference. We also include the results for SA-SSD reported in the original paper~\cite{he2020structure}, as these differ somewhat from those obtained with the provided pre-trained model. In Table~\ref{tab:kitti_test}, we observe that the added EBM and gradient-based refinement consistently improves the SA-SSD baseline across all metrics. We also observe that our SA-SSD+EBM detector achieves very competitive performance compared to previous methods. 

Results on KITTI \textit{val} in terms of 3D and BEV AP ($0.7$ threshold) are found in Table~\ref{tab:kitti_val}. There, we only include detectors for which AP values computed using 40 recall positions are available. In Table~\ref{tab:kitti_val}, we again observe that our EBM-based detector consistently outperforms the SA-SSD baseline. On KITTI \textit{val}, our SA-SSD+EBM also sets a new state-of-the-art in terms of all but one of the metrics. 

\begin{figure}[t]
    \centering
            \begin{tikzpicture}[scale=0.90]
                \pgfplotsset{
                    y axis style/.style={
                        yticklabel style=#1,
                        ylabel style=#1,
                        y axis line style=#1,
                        ytick style=#1
                  }
                }
            
                \begin{axis}[
                        axis y line*=left,
                        xlabel={Number of gradient ascent iterations $T$},
                        ylabel={Average AP},
                        xtick={0, 10, 20, 30, 40, 50, 60},
                        legend pos=north east,
                        grid style=dashed,
                        y tick label style={
                            /pgf/number format/.cd,
                                fixed,
                                fixed zerofill,
                                precision=1,
                            /tikz/.cd
                        },
                        every axis plot/.append style={thick},
                        y axis style=blue!65!black,
                    ]
                \addplot[smooth,mark=*,blue] 
                     plot [error bars/.cd, y dir = both, y explicit]
                     table[row sep=crcr, x index=0, y index=1]{
                    0 86.5666666667\\
                    1 86.71\\
                    2 87.3233333333\\
                    4 88.16\\
                    8 88.1966666667\\
                    10 88.17\\
                    16 88.1454\\
                    32 88.1193333333\\
                    64 88.1233333333\\
                    };
                \end{axis}
                
                \begin{axis}[
                    axis y line*=right,
                    axis x line=none,
                    xlabel={Number of gradient ascent iterations $T$},
                    ylabel={FPS},
                    legend pos=north east,
                    grid style=dashed,
                    y tick label style={
                        /pgf/number format/.cd,
                            fixed,
                            fixed zerofill,
                            precision=0,
                        /tikz/.cd
                    },
                    every axis plot/.append style={thick},
                    y axis style=red!65!black,
                ]
                ]
                \addplot[smooth,mark=*,red] 
                     plot [error bars/.cd, y dir = both, y explicit]
                     table[row sep=crcr, x index=0, y index=1]{
                    0 19.1668459797\\
                    1 16.8603231361\\
                    2 15.3684413143\\
                    4 12.8144810124\\
                    8 9.67558682366\\
                    10 8.41298310584\\
                    16 6.43987684987\\
                    32 3.76447974932\\
                    64 2.14341369469\\
                    };
                \end{axis}
            \end{tikzpicture}\vspace{0.0mm}
      \caption{Impact of the number of gradient ascent iterations $T$ in Algorithm~\ref{algo:prediction} on detector performance (3D AP with 0.7 threshold, averaged over easy, moderate and hard) and detector inference speed (FPS), on KITTI \textit{val}. Refinement with $T=4$ iterations significantly improves the detector performance, while only decreasing the inference speed from $19.2$ to $12.8$ FPS.}
      \label{fig:impact_of_grad_iters}
\end{figure}
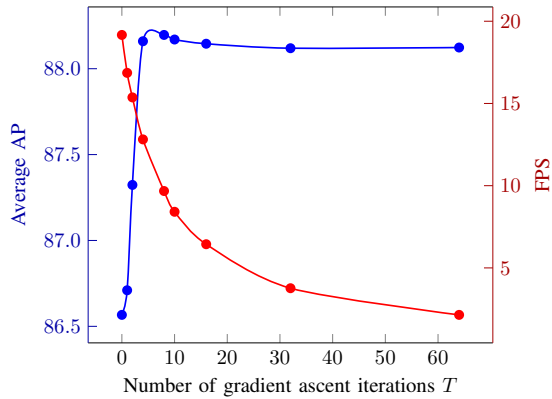

A further comparison of SA-SSD+EBM and the SA-SSD baseline is provided in Table~\ref{tab:kitti_val_3d}. There, we report AP for higher thresholds $\{0.75, 0.8, 0.85, 0.9\}$ on KITTI \textit{val}. We observe that the gradient-based refinement consistently improves detector performance across all metrics, and that the relative gain is larger for higher thresholds. Our approach thus also refines accurate bounding boxes even further, an effect not captured by the standard AP metrics.


\subsection{Analysis of Inference Speed}
The improved detection performance compared to SA-SSD comes with a decreased inference speed. On a single NVIDIA TITAN Xp GPU, SA-SSD runs at $19.2$ FPS, while SA-SSD+EBM runs at $8.4$ FPS for $T=10$ gradient ascent iterations. We further analyze how the choice of $T$ affects detector inference speed and performance in Figure~\ref{fig:impact_of_grad_iters}. The performance is here given in terms of 3D AP (0.7 threshold) averaged over the three difficulty levels (easy, moderate, hard), on KITTI \textit{val}. We observe that the choice $T=4$ provides approximately equal performance compared to $T=10$, while only decreasing the inference speed to $12.8$ FPS. This trade-off between detector performance and inference speed could potentially be further improved by using fewer grid points in our RoIAlign variant, or by using a more lightweight energy prediction network branch. Our approach could also be very well-suited for offboard 3DOD \cite{qi2021offboard}, where inference speed is less of a concern. Exploring these directions is left for future work.

\begin{figure}[t]
    \centering
            \begin{tikzpicture}[scale=0.90]
                \pgfplotsset{
                    y axis style/.style={
                        yticklabel style=#1,
                        ylabel style=#1,
                        y axis line style=#1,
                        ytick style=#1
                  }
                }
            
                \begin{axis}[
                        xlabel={$\Delta\phi$},
                        ylabel={$f_\theta(x, y)$},
                        xtick={0, 3.14, 6.28},
                        legend pos=north east,
                        grid style=dashed,
                        y tick label style={
                            /pgf/number format/.cd,
                                fixed,
                                fixed zerofill,
                                precision=1,
                            /tikz/.cd
                        },
                        every axis plot/.append style={thick},
                    ]
                \addplot[smooth,mark=] 
                     plot [error bars/.cd, y dir = both, y explicit]
                     table[row sep=crcr, x index=0, y index=1]{
                    0.000000 48.068439\\
                    0.062832 42.294743\\
                    0.125664 33.688911\\
                    0.188496 24.110680\\
                    0.251327 14.706524\\
                    0.314159 7.491511\\
                    0.376991 3.683142\\
                    0.439823 1.996514\\
                    0.502655 1.739263\\
                    0.565487 1.576064\\
                    0.628319 1.712095\\
                    0.691150 0.691976\\
                    0.753982 -0.036267\\
                    0.816814 -0.438010\\
                    0.879646 -0.613474\\
                    0.942478 -0.532532\\
                    1.005310 -0.657090\\
                    1.068142 -0.622545\\
                    1.130973 -0.466947\\
                    1.193805 -0.006368\\
                    1.256637 0.723749\\
                    1.319469 1.080728\\
                    1.382301 1.373230\\
                    1.445133 1.640777\\
                    1.507964 1.647005\\
                    1.570796 1.592185\\
                    1.633628 2.196492\\
                    1.696460 3.070720\\
                    1.759292 3.360998\\
                    1.822124 3.093468\\
                    1.884956 2.246062\\
                    1.947787 1.050786\\
                    2.010619 0.219700\\
                    2.073451 -0.389902\\
                    2.136283 -1.443546\\
                    2.199115 -2.172216\\
                    2.261947 -2.527925\\
                    2.324779 -3.022400\\
                    2.387610 -3.212375\\
                    2.450442 -2.909080\\
                    2.513274 -2.989322\\
                    2.576106 -2.807961\\
                    2.638938 -1.916174\\
                    2.701770 -0.589522\\
                    2.764602 1.828000\\
                    2.827433 5.977865\\
                    2.890265 11.626287\\
                    2.953097 19.174915\\
                    3.015929 26.542351\\
                    3.078761 32.508335\\
                    3.141593 34.989399\\
                    3.204425 33.015808\\
                    3.267256 27.996273\\
                    3.330088 21.507042\\
                    3.392920 14.869912\\
                    3.455752 9.214403\\
                    3.518584 5.244743\\
                    3.581416 2.503688\\
                    3.644247 0.403629\\
                    3.707079 -1.110806\\
                    3.769911 -1.656872\\
                    3.832743 -2.559255\\
                    3.895575 -2.844301\\
                    3.958407 -2.721656\\
                    4.021239 -2.178404\\
                    4.084070 -1.723201\\
                    4.146902 -1.103538\\
                    4.209734 -0.333984\\
                    4.272566 0.213631\\
                    4.335398 0.674516\\
                    4.398230 1.395605\\
                    4.461062 2.014120\\
                    4.523893 2.675209\\
                    4.586725 4.084327\\
                    4.649557 5.330992\\
                    4.712389 6.058853\\
                    4.775221 6.102948\\
                    4.838053 6.126039\\
                    4.900885 5.968259\\
                    4.963716 5.296932\\
                    5.026548 3.772353\\
                    5.089380 2.374893\\
                    5.152212 1.712681\\
                    5.215044 1.714821\\
                    5.277876 1.431682\\
                    5.340708 1.025762\\
                    5.403539 1.021235\\
                    5.466371 -0.101840\\
                    5.529203 -1.292487\\
                    5.592035 -2.034903\\
                    5.654867 -3.118789\\
                    5.717699 -4.276930\\
                    5.780530 -4.497327\\
                    5.843362 -4.232416\\
                    5.906194 -2.231362\\
                    5.969026 1.999620\\
                    6.031858 6.988460\\
                    6.094690 16.449743\\
                    6.157522 28.313625\\
                    6.220353 40.106926\\
                    6.283185 48.068401\\
                    };
                \end{axis}
            \end{tikzpicture}\vspace{0.0mm}
    \caption{Visualization of the DNN scalar output $f_\theta(x, y)$ when a predicted 3D bounding box $y$ (\ref{eq:y_def}) is rotated $\Delta\phi$ rad. The two distinct modes at $\Delta\phi= 0$ and $\Delta\phi = \pi$ demonstrate that the trained EBM $p(y|x;\theta)$ captures the inherent multi-modality in $p(y|x)$.}
    \label{fig:viz_angle}
\end{figure}
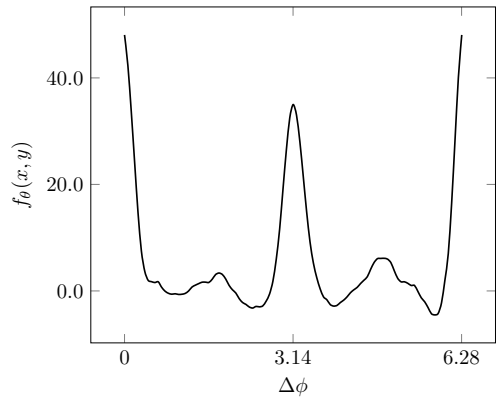

\subsection{Analysis of Learned Distribution}
For 3DOD from LiDAR point clouds, it can be inherently difficult to correctly predict the heading angle $\phi$ of a 3D bounding box $y$ (\ref{eq:y_def}). This is because it is often difficult, when only given a point cloud, to distinguish between two otherwise identical cars which are heading in opposite directions. The true distribution $p(y|x)$ will thus often have two distinct modes, one at the true heading angle $\phi$ and one at $\phi + \pi$. In Figure~\ref{fig:viz_angle}, we visualize $f_\theta(x, y) \in \mathbb{R}$ as a function of $\Delta\phi$ when a predicted 3D bounding box $y$ is rotated $\Delta\phi$ rad, demonstrating that our trained EBM $p(y|x;\theta)$ does indeed capture this inherent multi-modality in the true $p(y|x)$. Future directions include investigating if the trained EBM $p(y|x;\theta)$ could be used to construct accurate estimates of prediction uncertainty, or provide other useful insights.

\section{Conclusion}
\label{section:conclusion}

We applied conditional EBMs $p(y | x; \theta)$ to the task of 3D bounding box regression, thus extending the recent EBM-based regression approach from 2D to 3D object detection. By designing a differentiable pooling operator for 3D bounding boxes, we could integrate a conditional EBM $p(y | x; \theta)$ into the state-of-the-art 3D object detector SA-SSD. On the KITTI dataset, our approach consistently outperformed the SA-SSD baseline across all 3DOD metrics, and achieved highly competitive performance also compared to other state-of-the-art methods. By demonstrating the potential of EBM-based regression for highly accurate 3DOD, we hope that our work will encourage the research community to further explore the application of EBMs to 3DOD and other important regression tasks.

\parsection{Acknowledgments}
This research was supported by the Swedish Foundation for Strategic Research via the project \emph{ASSEMBLE}, the Knut and Alice Wallenberg Foundation via the \emph{Wallenberg AI, Autonomous Systems and Software Program (WASP)}, and the \emph{Kjell \& M\"arta Beijer Foundation}.

{\small
\bibliographystyle{ieee_fullname}
\bibliography{references}
}

\end{document}